\documentclass[a4paper,10pt,twocolumn]{article}

\usepackage[english]{babel}
\usepackage[utf8]{inputenc}
\usepackage[T1]{fontenc}

\usepackage[top=1.5cm, left=1.5cm, right=1.5cm, bottom=1.5cm]{geometry}

\renewenvironment{abstract}{\bf\small {\em\ Abstract---}}{}

\usepackage{amsfonts,amssymb,amsmath,amsthm}
\usepackage{subfigure}
\usepackage{graphicx}
\usepackage[footnotesize]{caption}

\usepackage{amsfonts,amssymb,amsmath,amsthm}
\usepackage{xspace}
\usepackage[ruled, onelanguage, linesnumbered]{algorithm2e}
\usepackage{comment}

\title{Frank-Wolfe Algorithm for the \ExactSparse Problem}

\author{Farah Cherfaoui$^{1}$\thanks{This work was supported by the Agence Nationale de la Recherche under grant JCJC MAD (ANR-14-CE27-0002).},  Valentin Emiya$^{1}$, Liva Ralaivola$^1$ and  Sandrine Anthoine$^2$\\
  \footnotesize $^1$ Aix Marseille Univ, Université de Toulon, CNRS, LIS, Marseille, France\\
\footnotesize $^2$ Aix Marseille Univ, CNRS, Centrale Marseille, I2M, Marseille, France} 
  
\date{\empty} 

\newtheorem{Theorem}{Theorem }
\newenvironment{sketch of proof}{%
  \proof}{\endproof}
\newtheorem{Lemma}{Lemma }

\newcommand{\convexe}{\mathcal{C}\xspace} 
\newcommand{\Boule}{\mathcal{B}\xspace} 
\newcommand{\ExactSparse}{m-\textsc{Exact-Sparse}\xspace}

\newcommand{\signal}{y\xspace} 
\newcommand{\argmax}{\operatorname{arg\,max}}
\newcommand{\argmin}{\operatorname{arg\,min}}
\newcommand{\sign}{\operatorname{sign}}

\newcommand{\conv}{\operatorname{conv}}
\newcommand{\support}{\operatorname{support}}

\newcommand{\norme}[1]{\| {#1} \|} 
\newcommand{\va}[1]{| {#1} |} 
\newcommand{\ps}[1]{\langle {#1} \rangle} 

\begin{document}

\maketitle

\begin{abstract} 
In this paper, we study the properties of the Frank-Wolfe algorithm to solve the \ExactSparse reconstruction problem. We prove that when the dictionary is quasi-incoherent, at each iteration, the Frank-Wolfe algorithm picks up an atom indexed by the support. We also prove that when the dictionary is quasi-incoherent, there exists an iteration beyond which the algorithm converges exponentially fast.
\end{abstract}

\section{Introduction} \label{sec:introduction}
Given a dictionary of a large number of atoms, the sparse signal approximation problem consists of constructing the best linear combination with a small number of atoms to approximate a given signal.
Sparse signal reconstruction is a sub-problem of the sparse signal approximation problem. In the latter case, we suppose that the given signal has an exact representation with $m$ or less atoms from this dictionary. We say that the signal is $m$-sparse. This subset of atoms is indexed by a set called the support. In this paper, we only consider the sparse signal reconstruction problem, which is called the \ExactSparse problem. 

Several algorithms have been developed to solve or approximate the \ExactSparse problem. The Matching Pursuit algorithm (MP) \cite{MZ92} and Orthogonal Matching Pursuit algorithm (OMP) \cite{PRK93} are two fundamental greedy algorithms used for solving this problem. Tropp~\cite{T04} and Gribonval and Vandergheynst~\cite{GV06} proved that, if the dictionary is quasi-incoherent, then at each iteration the MP and OMP algorithms pick up an atom indexed by the support. They also proved that these two algorithms converge exponentially fast. In fact, Tropp in~\cite{T04} demonstrates that OMP converges after exactly $m$ iterations, where $m$ is the size of the support. We study in this paper the properties of the Frank-Wolfe algorithm \cite{FW56} to solve the \ExactSparse problem.  
The Frank-Wolfe algorithm \cite{FW56} is an iterative optimization algorithm designed for constrained convex optimization. It has been proven to converge exponentially if the objective function is strongly convex \cite{GM86} and linearly in the other cases \cite{FW56}.
The atom selection steps in Matching Pursuit and Frank-Wolfe are very similar. This inspired for example Jaggi and al.~\cite{LKTJ17} to use the Frank-Wolfe algorithm to prove the convergence of the MP algorithm when no conditions are made on the dictionary.


In this paper, we use the MP algorithm to prove that the Frank-Wolfe algorithm can have the same recovery and convergence properties as MP. We prove that when the dictionary is quasi-incoherent, the Frank-Wolfe algorithm picks up only atoms indexed by the support. Also, we prove that when the dictionary is quasi-incoherent, the Frank-Wolfe algorithm converges exponentially from a certain iteration even though the function we consider is not strongly convex. 


\section{The problem and the algorithm}\label{section: FW for ExactSparse}
\subsection{The \ExactSparse problem}
For any vector $x \in \mathbb{R}^n$, we denote by $x(i)$ its $i^{th}$ coordinate. The support of $x$ is the set of indices of nonzero coefficients:
$$ \support(x) = \{i | x(i) \neq 0 \}. $$

Fix a dictionary $\Phi = [\varphi_1, \dots, \varphi_n] \in \mathbb{R}^{d \times n}$ of $n$ unit-norm vectors. Assume that $\signal$ is $m$-sparse, then the \ExactSparse problem is to find:
\begin{equation*}\label{pb exactSparse}
\begin{split} 
  \argmin_{x \in \mathbb{R}^n } \tfrac{1}{2}\norme{\signal - \Phi x}^2_2 \ \ \text{s.t. } \norme{ x }_0 \leq m 
 \end{split} 
\end{equation*}
where the $l_0$ pseudo-norm $\norme{.}_0$ counts the number of nonzero components in its argument. 
This problem has been proven to be NP-hard \cite{DMA97} and has been tackled essentially with two kind of approaches. The first one is the local approach, using a greedy algorithm like MP or OMP. The second approach is a global one where one relaxes the problem. A most popular choice is the $l_1$ relaxation:
\begin{equation}\label{equation: problem ExactSparse}
\begin{split} 
  \argmin_{x \in \mathbb{R}^n } \tfrac{1}{2}\norme{\signal - \Phi x}^2_2  \ \ \text{s.t. } \norme{ x }_1 \leq \beta 
 \end{split} 
\end{equation}
where $\norme{.}_1$ is the $l_1$ norm.

We present, in the next parts, the Frank-Wolfe algorithm \cite{FW56} for the \ExactSparse problem, and then the recovery properties and convergence rate of this algorithm.

\subsection{The Frank-Wolfe algorithm}
The Frank-Wolfe algorithm solves the optimization problem
\begin{equation*}\label{equation: optimization}
\begin{split} 
  \min_{x \in \convexe} f(x)  \ \ \text{s.t. } x \in \convexe 
 \end{split} 
\end{equation*}
where $f$ is a convex and continuously differentiable function and $\convexe$ is a compact and convex set. In the original version of the Frank-Wolfe algorithm, each iterate $x_{k+1}$ is defined as a convex combination between $x_k$ and $s_k$ 
with $s_k = \argmin_{s \in \convexe} \langle s,\nabla f(x_k)\rangle$.

In the case of the relaxation of the \ExactSparse problem (Equation~\eqref{equation: problem ExactSparse}), $f(x) = \frac{1}{2}\norme{\signal - \Phi x}^2_2$ and $\convexe = \{ x: \ \norme{x}_1 \leq \beta \} = \Boule_1(\beta)$ is the $l_1$ ball of radius $\beta$.
Noting that $\Boule_1(\beta) = \conv\{ \pm \beta e_i | i \in \{1,\dots,n \} \}$ and that $\nabla f(x) = \Phi^{t}(\Phi x - \signal)$, we obtain that $s_k$ can be calculated as in line 4 and 5 of Algorithm~\ref{algorithm: Frank-Wolfe}. Note also that we initialize $x_0$ by zero (line 1) and that we select the convex combination parameter $\gamma_k$ as in line 6.

\begin{algorithm}[t] 
 \KwData{signal $\signal$, dictionary $\Phi = [\varphi_1, \dots, \varphi_n]$, scalar $\beta$.}
 $x_0 = 0 $\\
 $k = 0$\\
 \While{stopping criterion not verified}{
 $i_{k} = \argmax_{i \in \{1,\dots,n \} } | \langle \varphi_i,\Phi x_k - \signal \rangle | $\\ 
 $s_k = -\sign(\langle \varphi_{i_k},\Phi x_k - \signal \rangle)\beta e_{i_{k}}$\\
 $\gamma_k = \argmin_{\gamma \in [0,1]} \norme{ \signal - \Phi (x_k + \gamma(s_k - x_k)) }_2^2$ \\
 $x_{k+1} = x_k + \gamma_k(s_k - x_k)$\\
 $k = k + 1$\\
 }
 \caption{Frank-Wolfe algorithm}
 \label{algorithm: Frank-Wolfe}
\end{algorithm}

In the analysis of Algorithm~\ref{algorithm: Frank-Wolfe}, we use the residual $r_k =  \signal - \Phi x_k$ whose norm is also the minimized objective function $f(x_k) = \frac{1}{2} \norme{r_k}_2^2 $.

\section{Recovery property and convergence rate}
\label{section: recovery and convergence}
For a dictionary $\Phi$, we denote by $\mu = \max\limits_{j \neq k} \va{ \langle \varphi_{j}, \varphi_{k} \rangle }$ the coherence of $\Phi$ and by $\mu_1(m) = \max\limits_{| \Lambda | = m}\max\limits_{i \notin \Lambda} \sum\limits_{j \in \Lambda} \va{ \ps{ \varphi_i, \varphi_{j} } }$ the Babel function. These two quantities measure how much the elements of the dictionary look alike. More details can be found in \cite{T04}.

In this section we present our major results. Theorem \ref{theorem: recovery property} gives the recovery property for the Frank-Wolfe algorithm. We prove that when the dictionary is quasi-incoherent (i.e. $m< \frac{1}{2}(\mu^{-1} + 1)$), the Frank-Wolfe algorithm reconstructs every $m$-sparse signal. Theorem \ref{theorem: convergence rate} shows that when the dictionary is quasi-incoherent, the Frank-Wolfe algorithm converges exponentially. We recall that a sequence $(a_k)_{k=0}^\infty$ converges exponentially if: $\forall \ k\in \{1,\dots,+\infty \}$, $a_{k+1} \leq qa_k$ with $0 < q < 1$.

\begin{Theorem}\label{theorem: recovery property}
Let $\Phi \in \mathbb{R}^{d\times n}$ be a dictionary, $\mu$ its coherence, and $y = \Phi x^*$ a $m$-sparse signal (i.e. $|\support(x^*) |= m$). \\
If $ m < \frac{1}{2}(\mu^{-1} + 1) $, then at each iteration, Algorithm \ref{algorithm: Frank-Wolfe} picks up a correct atom, i.e. $\forall \ k$, $ i_k \in \support(x^*)$.
\end{Theorem}

\begin{sketch of proof}
The proof of this theorem is very similar to the proof of Theorem 3.1 in \cite{T04}.
\end{sketch of proof}


\begin{Theorem}\label{theorem: convergence rate}
Let $\Phi \in \mathbb{R}^{d\times n}$ be a dictionary, $\mu$ its coherence, and $y = \Phi x^*$ a $m$-sparse signal (i.e. $|\support(x^*) |= m$). \\
If $ m < \frac{1}{2}(\mu^{-1} + 1) $ and $\norme{x^*}_1 < \beta$, then there exists a $K$ such that for all iteration $k \geq K$ of Algorithm \ref{algorithm: Frank-Wolfe}, we have:
$$ \| r_{k+1} \|^2 \leq \| r_{k} \|^2 \Bigg( 1 - \frac{\epsilon^2(1 - \mu_1(m-1))}{4\beta^2} \Bigg)$$
where $\epsilon = \frac{1}{2} (\beta - \norme{x^*}_1)$.
\end{Theorem}

\begin{sketch of proof}
The general idea of the proof can be summarized as follows. The first step will be to prove that if the dictionary is quasi-incoherent, then the step $\gamma_k$ chosen in line 6 of Algorithm \ref{algorithm: Frank-Wolfe} is in $(0, 1)$. A consequence of this is that:
\begin{align}
  \gamma_k & = \argmin_{\gamma \in \mathbb{R}} \norme{ \signal - \Phi (x_k + \gamma(s_k - x_k)) }_2^2 \\
\label{eq:gamma}
           & = \frac{\ps{ r_k, \Phi(s_k - x_k) }}{\norme{\Phi(s_k - x_k)}^2_2}
\end{align}
%
We can then write the expression of $\norme{r_{k+1}}_2^2$: 
$$ \norme{r_{k+1}}_2^2 = \norme{\signal - \Phi x_{k+1}}_2^2 = \norme{r_{k} + \gamma_k\Phi(s_k - x_k) }_2^2, $$
which yields using Eq.~\eqref{eq:gamma}:
$$ \norme{r_{k+1}}^2_2 = \norme{r_{k}}^2_2 - \frac{\ps{r_k, \Phi(s_k - x_k)}^2}{\norme{\Phi(s_k - x_k)}_2^2}. $$
The second step is to bound $\ps{r_k, \Phi(s_k - x_k)}$. Using Theorem \ref{theorem: recovery property}, we can show that the sequence of $\norme{x_k - x^*}$ is bounded by the sequence $f(x_k) - f(x^*)$. Since the sequence $f(x_k) - f(x^*)$ converges to zero, then the sequence of $\norme{x_k - x^*}$ also converges to zero. 
Therefore, there exists an iteration $K$ such that for all $k \geq K$: $x_k \in \Boule_2(x^*, \epsilon)$ where $\Boule_2(x^*, \epsilon)$ is $l_2$ ball centered in $x^*$ and of radius $\epsilon$. As a result, $x_k - \epsilon\frac{\nabla f(x_k)}{\norme{\nabla f(x_k)} } \in \Boule_2(x^*, 2\epsilon)$. 
Since $\norme{x^*}_1 + 2\epsilon < \beta$, we have $\Boule_2(x^*, 2\epsilon) \subseteq \Boule_1(\beta)$.

By definition of $s_k$:
$$ \ps{ s_k , \nabla f(x_k)} \leq \ps{ x_k - \epsilon \frac{\nabla f(x_k)}{\| \nabla f(x_k) \|} , \nabla f(x_k)}. $$
Noting that $\nabla f(x_k)=-\Phi^t r_k$, one obtains
$$ \ps{ r_k, \Phi(s_k - x_k) } \geq \epsilon \norme{\Phi^t r_k}. $$
By Theorem \ref{theorem: recovery property}, $r_k$ lies in the linear span of atoms indexed by $\support(x^*)$. Since we assume that these atoms are linearly independent, we have
$$ \norme{\Phi^t r_k} \geq \lambda_{min}^{\Phi_{ \support(x^*)}} \norme{r_k}_2,$$
where $\Phi_{ \support(x^*)}$ is the matrix whose columns are the atoms indexed by $\support(x^*)$ and $\lambda_{min}^{\Phi_{ \support(x^*)}}$ its smallest singular. So,
%
By Lemma 2.3 of~\cite{T04}, $\lambda_{min}^{\Phi_{ \support(x^*)}} \geq (1 - \mu_1(m-1))$ and we obtain:
$$\ps{ r_k, \Phi(s_k - x_k) }  \geq \epsilon (1 - \mu_1(m-1)) \norme{r_k}_2.$$

Finally, we show that $\norme{\Phi(s_k - x_k)}_2 \leq 2 \beta $ using the fact that $\norme{\Phi ((s_k - x_k))}_2\leq \norme{s_k - x_k}_1$ since the $\varphi_i$ are of unit norm. 
\end{sketch of proof}
Note that Tropp in~\cite{T04} has already proved that if the dictionary is incoherent, then $\mu_1(m) + \mu_1(m-1) < 1$. As a result, $1 - \mu_1(m-1)$ is in $(0, 1)$. We also have that $\frac{\epsilon^2}{\beta^2} < 1$ because $\epsilon < \beta$. Finally, since $d$ is greater that $1$, we have that $\tfrac{\epsilon^2(1 - \mu_1(m-1))}{4\beta^2d}$ is in $(0, 1)$. We conclude that Theorem \ref{theorem: convergence rate} gives the exponential convergence rate of the residual norm. As $f(x_k) = \frac{1}{2}\norme{r_k}^2_2$, this implies that this theorem also gives the exponential convergence rate of the objective function beyond a certain iteration.

It is possible to guarantee an exponential convergence from the first iteration if $\beta$ is big enough. Lemma \ref{lemma: min value for beta} gives a lower bound of $\beta$ to obtain this result.

\begin{Lemma}\label{lemma: min value for beta}
Let $\Phi$ be a dictionary of coherence $\mu$, $y = \Phi x^*$ a $m$-sparse signal (i.e. $|\support(x^*) |= m$) and $\epsilon \in (0, 1)$. If 
$$\beta > \frac{m \norme{\signal}_2}{\epsilon\lambda^{ \Phi_{ \support(x^*)} }_{min}} \Bigg( 1 + \frac{\lambda^{\Phi_{ \support(x^*)}}_{max} }{ \lambda^{\Phi_{ \support(x^*)}}_{min} } \Bigg)$$
then Algorithm \ref{algorithm: Frank-Wolfe} converges exponentially from the first iteration.
Here, $\Phi_{ \support(x^*)}$ is the matrix whose columns are the atoms indexed by $\support(x^*)$.
\end{Lemma}

We proved in Theorem \ref{theorem: convergence rate} that when the iterates $x_k$ enter the ball $\Boule_1(x^*, \epsilon)$, the Frank-Wolfe algorithm converges exponentially. The intuition of this lemma is to grow the value of $\beta$ compared to $\norme{x^*}$ (then $\epsilon$ also grows). This implies that the iterates $x_k$ enter the ball $\Boule_1(x^*, \epsilon)$ earlier and the exponential convergence starts earlier.

\bibliographystyle{plain}
\bibliography{biblio}

\begin{thebibliography}{1}

\bibitem{DMA97}
Geoff Davis, Stephane Mallat, and Marco Avellaneda.
\newblock Adaptive greedy approximations.
\newblock {\em Constructive approximation}, 13(1):57--98, 1997.

\bibitem{FW56}
Marguerite Frank and Philip Wolfe.
\newblock An algorithm for quadratic programming.
\newblock {\em Naval Research Logistics (NRL)}, 3(1-2):95--110, 1956.

\bibitem{GV06}
R{\'e}mi Gribonval and Pierre Vandergheynst.
\newblock On the exponential convergence of matching pursuits in
  quasi-incoherent dictionaries.
\newblock {\em IEEE Transactions on Information Theory}, 52(1):255--261, 2006.

\bibitem{GM86}
Jacques Gu{\'e}lat and Patrice Marcotte.
\newblock Some comments on {W}olfe's ‘away step’.
\newblock {\em Mathematical Programming}, 35(1):110--119, 1986.

\bibitem{LKTJ17}
Francesco Locatello, Rajiv Khanna, Michael Tschannen, and Martin Jaggi.
\newblock A unified optimization view on generalized matching pursuit and
  {Frank-Wolfe}.
\newblock {\em arXiv preprint arXiv:1702.06457}, 2017.

\bibitem{MZ92}
St{\'e}phane~G Mallat and Zhifeng Zhang.
\newblock Matching pursuits with time-frequency dictionaries.
\newblock {\em IEEE Transactions on signal processing}, 41(12):3397--3415,
  1993.

\bibitem{PRK93}
Yagyensh~Chandra Pati, Ramin Rezaiifar, and Perinkulam~Sambamurthy
  Krishnaprasad.
\newblock Orthogonal matching pursuit: Recursive function approximation with
  applications to wavelet decomposition.
\newblock {\em Signals, Systems and Computers, 1993. 1993 Conference Record of
  The Twenty-Seventh Asilomar Conference on}, pages 40--44, 1993.

\bibitem{T04}
Joel~A Tropp.
\newblock Greed is good: Algorithmic results for sparse approximation.
\newblock {\em IEEE Transactions on Information theory}, 50(10):2231--2242,
  2004.

\end{thebibliography}


\begin{thebibliography}{10}
\bibitem{T04}
Tropp Joel A, 
\newblock ``Greed is good: Algorithmic results for sparse approximation'',
\newblock IEEE Transactions on Information theory, {\bf 50}(10):2231--2242, 2004

\bibitem{GV06}
Gribonval R{\'e}mi and Vandergheynst Pierre,
\newblock ``On the exponential convergence of matching pursuits in quasi-incoherent dictionaries'',
\newblock IEEE Transactions on Information Theory, {\bf 52}(1):255--261, 2006

\bibitem{FW56}
Frank Marguerite and Wolfe Philip,
\newblock ``An algorithm for quadratic programming'',
\newblock Naval Research Logistics (NRL), {\bf 3}(1-2):95--110, 1956

\bibitem{DMA97}
Davis Geoff and Mallat Stephane and Avellaneda Marco,
\newblock ``Adaptive greedy approximations'',
\newblock Constructive approximation, {\bf 13}(1):57--98, 1997

\bibitem{MZ92}
Mallat St{\'e}phane G and Zhang Zhifeng,
\newblock `` Matching pursuits with time-frequency dictionaries'',
\newblock IEEE Transactions on signal processing, {\bf 41}(12):3397--3415, 1993

\bibitem{PRK93}
Pati Yagyensh Chandra and Rezaiifar Ramin and Krishnaprasad Perinkulam Sambamurthy,
\newblock ``Orthogonal matching pursuit: Recursive function approximation with applications to wavelet decomposition '',
\newblock Signals, Systems and Computers, 1993

\end{thebibliography}

\end{document}